\documentclass{article} %
\usepackage{colm2024_conference}

\usepackage{booktabs}
\usepackage{graphicx}
\usepackage{enumitem}
\usepackage{wrapfig}
\usepackage{algorithm}
\usepackage{algpseudocode}

\usepackage{microtype}
\usepackage{amsmath}
\usepackage{colortbl}
\usepackage[utf8]{inputenc}
\definecolor{lightgray}{rgb}{0.9,0.9,0.9}
\usepackage{caption}
\usepackage{subcaption}
\usepackage{xcolor}
\usepackage{setspace}
\usepackage{url}
\usepackage{multirow}
\usepackage{colortbl}
\usepackage{tabularx}
\usepackage{blindtext}
\usepackage{pgfplots}
\pgfplotsset{compat=1.18} 
\usepackage{tikz}
\usetikzlibrary{er,positioning,bayesnet}
\usepackage{makecell}
\usepackage{tipa}
\usepackage{siunitx}
\usepackage{nicefrac}
\usepackage{tocloft}
\usepackage{listings}
\usepackage[raster,skins]{tcolorbox} %
\usepackage{xltabular}
\usepackage{adjustbox}
\usepackage{xurl}
\usepackage{cleveref}

\usepackage{amsmath,amsfonts,bm}

\def\eqref#1{equation~\ref{#1}}

\def\1{\bm{1}}

\DeclareMathAlphabet{\mathsfit}{\encodingdefault}{\sfdefault}{m}{sl}
\SetMathAlphabet{\mathsfit}{bold}{\encodingdefault}{\sfdefault}{bx}{n}

\definecolor{codegreen}{rgb}{0,0.6,0}
\definecolor{codegray}{rgb}{0.5,0.5,0.5}
\definecolor{codepurple}{rgb}{0.58,0,0.82}
\definecolor{backcolour}{rgb}{0.95,0.95,0.92}
\definecolor{boxblue}{RGB}{57,89,163}
\definecolor{boxbluebg}{RGB}{230,237,250} 
\definecolor{mygray}{gray}{0.7}
\definecolor{upcolor}{RGB}{57,182,74}
\newcommand{\up}[1]{\textcolor{upcolor}{$\uparrow$ #1}}

\title{Guiding the Inner Eye: A Framework for Hierarchical and Flexible Visual Grounded Reasoning}

\author{
  \bf Zhaoyang Wei, Wenchao Ding, Yanchao Hao, Xi Chen\\
  Basic Algorithm Center, PCG, Tencent
}

\begin{document}

\maketitle

\begin{abstract}
Models capable of "thinking with images" by dynamically grounding their reasoning in visual evidence represent a major leap in multimodal AI. However, replicating and advancing this ability is non-trivial, with current methods often trapped between the instability of end-to-end reinforcement learning (RL) and the rigidity of supervised fine-tuning (SFT). This leads to models that either struggle to learn or lack the cognitive flexibility required for complex, real-world scenes. To navigate this dilemma, we introduce \textbf{GRiP} (Guided Reasoning and Perception), a novel two-stage training framework that cultivates robust and flexible visual grounded reasoning by explicitly guiding the model's perceptual focus and logical pathways. GRiP's core lies in its cognitive-enhanced RL stage, which features two key innovations: (1) a \textbf{Salience-Weighted IoU Reward} that incentivizes the model to prioritize the localization of mission-critical objects over trivial distractors, and (2) a \textbf{Multi-Heuristic Reward} that encourages cognitive flexibility by rewarding diverse yet logically valid reasoning pathways. Initialized from the Qwen2.5-VL-7B model, GRiP demonstrates significant performance gains across multiple challenging benchmarks. It achieves state-of-the-art results among open-source models on the highly challenging TreeBench and V* Bench, proving its effectiveness in complex visual reasoning. Our work demonstrates that moving beyond simplistic rewards and instead guiding models with cognitively-inspired signals for \textit{what} to see and \textit{how} to think is crucial for unlocking the next level of multimodal intelligence. The code will be made publicly available.
\end{abstract}

\vspace{-15pt}



\section{Introduction}
\label{sec:intro}

Recent breakthroughs in Vision-Language Models (VLMs) have pushed the frontiers of multimodal reasoning, largely by adopting long Chain-of-Thought (CoT) prompting~\citep{wei2022chain, team2025kimi}. Despite their success on various benchmarks~\citep{lu2023mathvista, yue2024mmmu}, these models predominantly operate within a text-centric paradigm. Their reasoning process, though complex, remains confined to the language modality, treating visual input as a static, one-off information source rather than a dynamic workspace. This paradigm falters on visually-intensive tasks where fine-grained details—such as subtle spatial relationships, tiny objects, or occluded states—are decisive~\citep{wang2025traceable}.

In stark contrast, human cognition seamlessly integrates vision and thought, actively "thinking with images" by performing sequential visual fixations to gather evidence and refine hypotheses. A recent milestone, the OpenAI o3 model~\citep{o3}, has demonstrated a similar capability, dynamically grounding its reasoning in visual regions through an interleaved Multimodal Chain-of-Thought (iMCoT). This represents a significant leap towards true multimodal intelligence, yet its underlying mechanisms remain undisclosed, posing a substantial challenge for the open-source community to replicate and build upon.

Inspired by this, recent works like DeepEyes~\citep{zheng2025deepeyes} and Pixel-Reasoner~\citep{su2025pixelreasoner} have explored methods to instill this "thinking with images" ability. However, cultivating this sophisticated skill presents a series of profound challenges:

\begin{itemize}[left=0pt]
    \item \textbf{The Reinforcement Learning (RL) Dilemma:} Training a model with end-to-end RL from scratch is fraught with difficulty. As observed by~\citet{zheng2025deepeyes}, a naive model is initially "reluctant" to use visual grounding tools. Early, random explorations often yield poor localizations and low rewards, leading to training instability and making it difficult for the desired behavior to emerge organically.

    \item \textbf{The Imitation Trap:} To circumvent the RL dilemma, a common strategy is to first use Supervised Fine-Tuning (SFT) or instruction tuning to bootstrap the model~\citep{su2025pixelreasoner}. While this teaches the model the basic syntax of grounded reasoning, it introduces a new problem: the "imitation trap." The model learns to rigidly mimic a single "golden" reasoning path from the training data, lacking the flexibility to adapt or explore alternative strategies. This brittleness becomes a critical failure point when faced with novel scenarios.

    \item \textbf{The Coarse Reward Problem:} Even when a model learns to ground its reasoning, existing reward mechanisms are often too coarse to foster deep, human-like cognition. Standard IoU-based rewards, for instance, are "flat"—they treat a critical piece of evidence and a trivial background object with equal importance. Furthermore, they typically reward similarity to a single reference trajectory, inadvertently penalizing creative yet valid alternative lines of thought. This leads to models that can "see" but cannot truly "reason" with nuance and flexibility.
\end{itemize}

To systematically address these challenges, we introduce \textbf{GRiP} (\textbf{G}uided \textbf{R}easoning and \textbf{P}erception), a novel two-stage training framework designed to cultivate robust and cognitively flexible visual grounded reasoning. GRiP navigates the complex trade-offs between exploration and exploitation, imitation and innovation, through a carefully designed training pipeline.

First, we acknowledge the necessity of a warm start and employ a \textbf{bootstrapping stage} with structured instruction tuning. This efficiently teaches the model the fundamental grammar of grounded reasoning, providing a stable foundation. Then, to escape the imitation trap and refine the model's policy, we introduce a \textbf{policy refinement stage} powered by a multi-faceted RL reward architecture. This architecture is our core contribution and features two key innovations designed to solve the coarse reward problem:
\begin{enumerate}
    \item A \textbf{Salience-Weighted IoU Reward ($\mathcal{R}_{sw\text{-IoU}}$)} that introduces a hierarchy into the visual space, explicitly rewarding the model for prioritizing the localization of mission-critical objects over trivial ones.
    \item A \textbf{Multi-Heuristic Reward ($\mathcal{R}_{\text{MHR}}$)} that rewards the model's reasoning trajectory based on its similarity to the best of several distinct, pre-defined cognitive pathways (e.g., bottom-up, top-down). This liberates the model from mimicking a single path and encourages cognitive flexibility.
\end{enumerate}

As shown in Figure~\ref{fig:pipeline}, our GRiP framework synergistically combines a stable SFT foundation with a sophisticated, cognitively-informed RL refinement process. Experimental results demonstrate that GRiP, built upon Qwen2.5-VL-7B~\cite{bai2025qwen25vl}, achieves state-of-the-art performance among open-source models on challenging benchmarks like TreeBench~\cite{wang2025traceable} and V*~\citep{wu2024vstar}, validating the effectiveness of our approach.

Our main contributions are summarized as follows:
\begin{itemize}[left=0pt]
    \item We identify and analyze the "Coarse Reward Problem" as a key bottleneck in developing advanced visual grounded reasoning, complementing existing insights on the RL dilemma and imitation trap.
    \item We propose GRiP, a two-stage training framework that systematically addresses these challenges by combining a stable SFT bootstrap with a novel, multi-faceted RL reward architecture.
    \item We introduce two innovative reward components: a Salience-Weighted IoU Reward to guide precise, hierarchical localization, and a Multi-Heuristic Reward to foster cognitive flexibility in reasoning.
\end{itemize}

\section{GRiP: A Guided Reasoning and Perception Framework}
\label{sec:method}

\begin{figure}[h]
    \centering
    \includegraphics[width=0.85\linewidth]{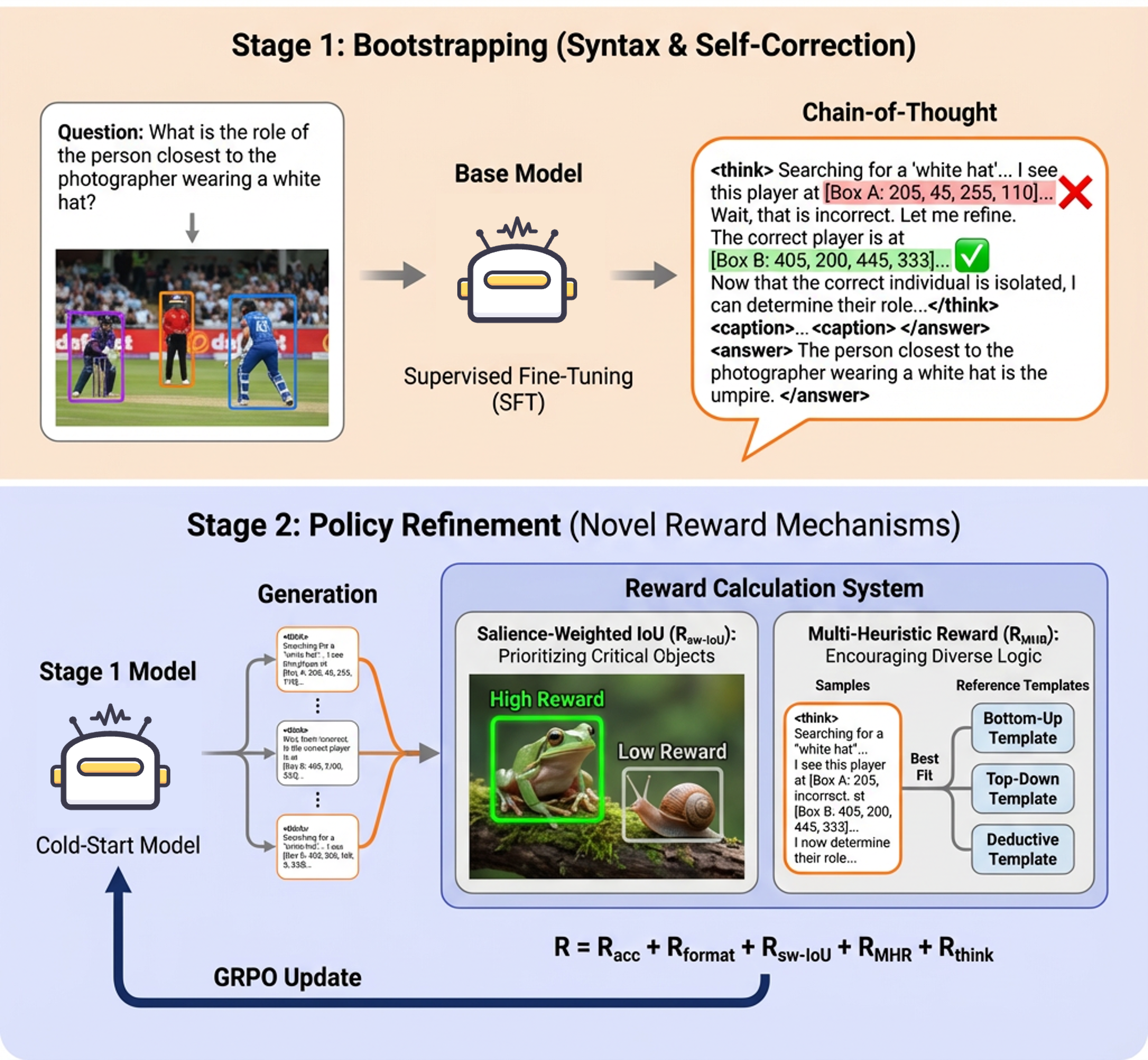} 
    \caption{The two-stage training pipeline of GRiP. (a) The bootstrapping stage uses supervised instruction tuning to teach the model the syntax of grounded reasoning. (b) The policy refinement stage employs reinforcement learning with our novel multi-faceted reward function, featuring the Salience-Weighted IoU Reward ($\mathcal{R}_{sw\text{-IoU}}$) and the Multi-Heuristic Reward ($\mathcal{R}_{\text{MHR}}$).}
    \label{fig:pipeline}
\end{figure}

Despite their progress, Multimodal Large Models (MLLMs) often exhibit brittleness in complex visual scenes, a failure largely attributable to a schism between their perceptual capabilities and reasoning processes. To remedy this, we propose \textbf{GRiP} (Guided Reasoning and Perception), a novel training framework designed to enforce a disciplined, human-like reasoning paradigm. The core principle of GRiP is to ensure that all reasoning is explicitly anchored to traceable visual evidence, creating a verifiable "perception-to-conclusion" chain. This principle is realized through a two-stage training architecture, as depicted in Figure~\ref{fig:pipeline}: (1) a bootstrapping stage using structured instruction tuning to establish foundational skills, and (2) a policy refinement stage using reinforcement learning with multi-faceted rewards to hone these skills.

\subsection{Stage 1: Bootstrapping with Structured Instruction Tuning}
\label{sec:bootstrap}

Deploying reinforcement learning (RL) on a naive model for a task as complex as visual grounded reasoning is impractical. The model would struggle in an astronomically large policy space, making the discovery of a valid reasoning structure highly improbable. To overcome this, our first stage focuses on \textbf{bootstrapping the model with structured instruction tuning}. The objective here is not to master the task, but to efficiently teach the model the fundamental \textit{syntax} of grounded reasoning: how to articulate a step-by-step analysis that correctly interleaves textual logic with spatial coordinates.

\paragraph{Instruction Tuning Data.} We constructed a dedicated dataset for this stage, \texttt{GRiP-SFT-35K}, containing 35k high-quality samples. We began by sampling high-resolution images from \texttt{SA-1B}~\cite{kirillov2023segment}, focusing on scenes rich with contextual complexity. For each image, our expert team created a question-answer pair along with a complete, "golden" reasoning trajectory. These trajectories were designed to model an explicit "localization-then-analysis" workflow. Crucially, to build in robustness, we also included a subset of samples designed to teach meta-cognition, where an incorrect bounding box is intentionally provided and immediately followed by a self-correction prompt (e.g., ``Wait, that localization is imprecise. Re-evaluating...''). This process yields a bootstrapped model, \textbf{GRiP-SFT}, which understands the required output format and provides a strong starting point for policy refinement.

\subsection{Stage 2: Policy Refinement with Multi-Faceted Rewards}
\label{sec:refine}

While instruction tuning provides a structural foundation, it is inherently rigid. To cultivate true reasoning flexibility and precision, we introduce a second stage of \textbf{policy refinement via reinforcement learning}. This stage moves beyond simple imitation and optimizes the model's policy using a sophisticated, multi-faceted reward signal that evaluates not only the final answer but also the quality and methodology of the reasoning process itself.

\subsubsection{Reward Architecture}
Our total reward, $R_{\text{total}}$, is a composite of four distinct signals: a standard accuracy reward ($R_{\text{acc}}$), a format compliance reward ($R_{\text{fmt}}$), and our two primary contributions: a \textbf{Salience-Weighted IoU Reward ($\mathcal{R}_{sw\text{-IoU}}$)} and a \textbf{Multi-Heuristic Reward ($\mathcal{R}_{\text{MHR}}$)}.
\begin{equation}
R_{\text{total}} = R_{\text{acc}} + R_{\text{fmt}} + \mathcal{R}_{sw\text{-IoU}} + \mathcal{R}_{\text{MHR}}
\end{equation}

\paragraph{Salience-Weighted IoU Reward ($\mathcal{R}_{sw\text{-IoU}}$).} In any given scene, some objects are more important than others for answering a specific question. Standard IoU rewards are agnostic to this fact, treating a distractor object and a critical piece of evidence with equal importance. To address this, we introduce the $\mathcal{R}_{sw\text{-IoU}}$, which modifies the dual IoU concept~\cite{wang2025traceable} by incorporating object salience.

We assign a salience score, $s_k > 1$, to mission-critical ground-truth objects ($g_k$), while non-essential objects retain a score of $s_k=1$. This score amplifies the reward for correctly localizing the most important evidence. The reward's \textbf{recall} component is thus weighted by salience:
\begin{equation}
\mathcal{R}_{\text{recall}} = \frac{1}{\sum s_k} \sum_{k=1}^{M} s_k \cdot \max_{i} \text{IoU}(p_i, g_k)
\end{equation}
where $\{p_i\}_{i=1}^N$ are the predicted boxes. This encourages the model to find all \textit{important} objects. The \textbf{precision} component remains unweighted to uniformly penalize any extraneous or "hallucinated" predictions, as all such errors are equally undesirable.
\begin{equation}
\mathcal{R}_{\text{precision}} = \frac{1}{N} \sum_{i=1}^{N} \max_{k} \text{IoU}(p_i, g_k)
\end{equation}
The final reward, $\mathcal{R}_{sw\text{-IoU}} = \frac{1}{2} (\mathcal{R}_{\text{recall}} + \mathcal{R}_{\text{precision}})$, thus guides the model toward a localization strategy that is both comprehensive and focused.

\paragraph{Multi-Heuristic Reward ($\mathcal{R}_{\text{MHR}}$).} Human cognition is flexible; we can solve problems using different lines of thought. To prevent the model from developing a brittle, single-track mind, we designed the $\mathcal{R}_{\text{MHR}}$ to reward diverse yet valid reasoning heuristics. We define three canonical heuristics:
\begin{enumerate}
    \item \textbf{Evidence-Driven (Bottom-Up):} Reasoning from specific visual details to a general conclusion.
    \item \textbf{Context-Aware (Top-Down):} Using scene-level understanding and common sense to interpret details.
    \item \textbf{Deductive Verification:} Treating the query as a hypothesis and systematically seeking confirmatory or contradictory evidence.
\end{enumerate}
For each training sample, we generate three reference trajectories ($\{\tau_{\text{ref}}^j\}_{j=1}^3$), one for each heuristic. The model's generated trajectory, $\tau_{\text{gen}}$, is rewarded based on its similarity to the \textit{best-matching} reference trajectory.
\begin{equation}
\mathcal{R}_{\text{MHR}} = \max_{j \in \{1,2,3\}} \text{sim}(\tau_{\text{gen}}, \tau_{\text{ref}}^j)
\end{equation}
This "best-of-three" approach encourages cognitive flexibility, teaching the model that there are multiple valid paths to a correct answer.

\subsubsection{Dataset for Policy Refinement}
\label{sec:rl_data}

This RL stage is powered by our \texttt{GRiP-RL-37K} dataset, a collection of 37k samples meticulously prepared for our multi-faceted reward system. We curated this dataset by first selecting hard samples from the \texttt{V*} benchmark~\cite{wu2024vstar} (i.e., those our \texttt{GRiP-SFT} model failed on) and then augmenting them with samples from other public datasets that feature complex relational reasoning. For every sample, our expert team performed a two-fold annotation process: (1) assigning salience scores ($s_k$) to all relevant objects to enable $\mathcal{R}_{sw\text{-IoU}}$, and (2) generating and manually refining three distinct reference trajectories ($\tau_{\text{ref}}^j$) to enable $\mathcal{R}_{\text{MHR}}$. This richly annotated dataset is the key to effectively training a model with our sophisticated reward signals, enabling it to learn not just \textit{what} the right answer is, but \textit{how} to arrive at it through precise, flexible, and human-like reasoning.

\subsection{Training Recipe}

Our model, \textbf{GRiP}, is initialized from the \texttt{Qwen2.5-VL-7B-Instruct}~\cite{team2024qwen25} checkpoint. The entire training process is conducted on 8 NVIDIA H20 (96GB) GPUs and follows the two-stage architecture detailed in Section~\ref{sec:method}: (1) Bootstrapping with Structured Instruction Tuning, and (2) Policy Refinement with Reinforcement Learning.

\paragraph{Stage 1: Bootstrapping.}
The initial SFT stage is performed on our \texttt{GRiP-SFT-35K} dataset. We utilize the \texttt{LLaMA-Factory}~\citep{zheng2024llamafactory} framework for efficient training. The optimization is carried out using the AdamW optimizer with a global batch size of 256. We employ a cosine learning rate schedule~\citep{loshchilov2016sgdr} with an initial learning rate of $5 \times 10^{-6}$ and a warmup ratio of 0.1. This stage provides the model with the foundational ability to generate structurally correct, grounded reasoning traces.

\paragraph{Stage 2: Policy Refinement.}
Starting from the \texttt{GRiP-SFT} checkpoint, we proceed to the RL stage using our \texttt{GRiP-RL-37K} dataset. A key challenge during this phase is \textit{policy degradation}, where the model, in pursuit of maximizing the recall component of the IoU reward, may learn to generate an excessive number of bounding boxes, leading to verbose and nonsensical outputs. Our reward architecture is designed to counteract this: the unweighted precision term in $\mathcal{R}_{sw\text{-IoU}}$ inherently penalizes such "hallucinated" boxes. To further stabilize training, we introduce a soft length penalty, $R_{\text{len}}$, which applies a small negative reward if the generated trajectory exceeds a predefined token limit. This discourages degenerate, overly long responses.

For the RL algorithm, we adopt Group Relative Policy Optimization (GRPO)~\citep{shao2024deepseekmath}, which we found to be more stable and effective for this task compared to vanilla PPO. The entire RL pipeline is implemented using the \texttt{EasyRL}~\cite{zheng2025easyr1} framework. The reward calculation, particularly the semantic similarity for $\mathcal{R}_{\text{MHR}}$, requires a powerful teacher model. For this role, we use \texttt{Qwen2.5-72B-Instruct}~\cite{team2024qwen25}, served via the \texttt{vLLM}~\cite{kwon2023vllm} inference engine on a separate set of 8 GPUs to ensure a high-throughput RL loop. The model is trained for 5 epochs on the RL dataset, which is sufficient for the policy to converge to a state

\begin{table}[t]
    \centering\small
    \setlength{\tabcolsep}{3pt}
    \resizebox{1.0\textwidth}{!}
    {\begin{tabular}{l ccccccccccccc}
    \toprule
    & & & \rotatebox{60}{Attributes} & \rotatebox{60}{Material} & \rotatebox{60}{Phy. State} & \rotatebox{60}{Obj. Retr.} & \rotatebox{60}{OCR} & \rotatebox{60}{Per. Trans.} & \rotatebox{60}{Ordering} & \rotatebox{60}{Con. \& Oc.} & \rotatebox{60}{Spa. Cont.} & \rotatebox{60}{Comparison} \\
    \cmidrule(lr){4-8}
    \cmidrule(lr){9-13}
    & \textbf{Overall} & \textbf{mIoU} & \multicolumn{5}{c}{Perception} & \multicolumn{5}{c}{Reasoning} \\
    \midrule
    \multicolumn{13}{c}{\textbf{Private Models}} \\
    \midrule
    Gemini-2.5-Flash-0520~\cite{gemini-2.5-flash} & 45.9 & -- & 48.3 & 53.9 & 69.6 & 68.8 & 75.0 & 15.3 & 19.3 & 56.1 & 72.4 & 43.2 \\
    GPT-4o-1120~\cite{gpt4o} & 46.9 & -- & 51.7 & 61.5 & 65.2 & 43.8 & 69.1 & 18.8 & 38.6 & 48.8 & 72.4 & 43.2 \\
    Gemini-2.5-Pro-0605~\cite{gemini-2.5-pro} & 54.1 & -- & 51.7 & 61.5 & 56.5 & 75.0 & 83.8 & 20.0 & 36.8 & 65.9 & 86.2 & 54.6 \\
    o3-0416~\cite{o3} & 54.8 & --$^\dag$ & 69.0 & 69.2 & 65.2 & 68.8 & 79.4 & 22.4 & 38.6 & 61.0 & 86.2 & 50.0 \\ 
    \midrule
    \multicolumn{13}{c}{\textbf{Open-source General Models}} \\
    \midrule
    LLaVA-OneVision-7B~\cite{li2024llavaov} & 37.3 & -- & 55.2 & 53.8 & 56.5 & 50.0 & 32.4 & 21.2 & 22.8 & 41.5 & 72.4 & 36.4 \\
    LLaVA-OneVision-72B~\cite{li2024llavaov} & 40.5 & -- & 62.1 & 53.8 & 65.2 & 62.3 & 36.8 & 12.9 & 28.1 & 53.7 & 65.5 & \textbf{47.7} \\
    MiniCPM-V-2.6-8B~\cite{yao2024minicpm} & 30.1 & -- & 51.7 & 46.2 & 60.9 & 62.5 & 42.6 & 0.0 & 3.5 & 36.6 & 62.1 & 29.5 \\
    Qwen2.5-VL-7B~\cite{bai2025qwen25vl} & 37.0 & -- & 55.2 & 53.8 & 56.5 & 62.5 & 27.9 & 20.0 & 35.1 & 39.0 & 44.8 & 43.2 \\
    Qwen2.5-VL-32B~\cite{bai2025qwen25vl} & 42.5 & -- & 51.7 & 53.8 & 69.6 & 62.5 & 54.4 & 16.5 & 33.3 & 46.3 & 62.1 & 38.6 \\
    Qwen2.5-VL-72B~\cite{bai2025qwen25vl} & 42.2 & -- & 65.5 & \textbf{69.2} & 56.5 & 56.3 & 48.5 & 11.8 & 33.3 & 51.2 & 72.4 & 38.6 \\
    InternVL3-8B~\cite{zhu2025internvl3} & 38.8 & -- & 51.7 & \textbf{69.2} & 56.5 & 56.3 & 33.7 & 21.2 & 24.6 & 39.0 & 72.4 & 43.2 \\
    InternVL3-38B~\cite{zhu2025internvl3} & 42.0 & -- & 51.7 & 61.5 & 52.2 & \textbf{68.8} & 51.5 & 12.9 & 33.3 & 56.1 & 65.5 & 38.6 \\
    InternVL3-78B~\cite{zhu2025internvl3} & 46.4 & -- & 62.1 & 61.5 & 52.2 & \textbf{68.8} & 52.9 & 16.5 & 33.3 & 61.0 & \textbf{86.2} & 45.5 \\
    \midrule
    \multicolumn{13}{c}{\textbf{Open-source Visual Grounded Reasoning Models}} \\
    \midrule
    DeepEyes-7B~\cite{zheng2025deepeyes} & 37.5 & 30.0 & 62.1 & 53.8 & 65.2 & 68.8 &  51.5 & 11.8 & 24.6 & 36.6 & 51.7 & \textbf{47.7} \\
    Pixel-Reasoner-7B~\cite{su2025pixelreasoner} & 39.0 & 35.7 & 58.6 & 61.5 & 65.2 & 50.0 & 48.5 & 14.1 & 31.6 & 39.0 & 44.8 & 40.9 \\
    TreeVGR-7B~\cite{wang2025traceable} & 50.4 & 44.0 & 65.5 & 53.8 & 82.6 & 68.8 & 63.3 & 22.4 & 36.8 & 61.0 & 69.0 & 45.5 \\
    \midrule
    \textbf{GRiP} & \textbf{51.3} & \textbf{45.0} & \textbf{69.1} & 54.4 & \textbf{83.4} & \textbf{69.3} & \textbf{64.1} & \textbf{23.2} & \textbf{38.6} & 58.7 & 69.4 & 45.8 \\
    $\Delta$ \textit{v.s.} Qwen2.5-VL-7B & \up{14.3} & -- & \up{13.9} & \up{0.6} & \up{26.9} & \up{6.8} & \up{36.2} & \up{3.2} & \up{3.5} & \up{19.7} & \up{24.6} & \up{2.6} \\
    \bottomrule
    \end{tabular}}
    \vspace{-5pt}
    \caption{
    Selected results of different models on treebench.
    Evaluations of open-source general models are implemented using VLMEvalKit~\cite{duan2024vlmevalkit}, while evaluations of visual grounded reasoning models are conducted by us.
    $^\dag$Reasoning pathways of o3~\cite{o3} are unavailable, and thus traceable evaluations are \textit{not} valid.
    Best performances for open-source models are highlighted in \textbf{bold}. \textit{Our} \textbf{GRiP} achieves comparable performance with InternVL3-78B~\cite{zhu2025internvl3}. }
    \label{tab:bench_table}
\end{table}
\begin{table}[t]
    \centering\small
    \begin{tabular}{l ccc ccc ccc}
    \toprule
    & \multicolumn{3}{c}{V* Bench} & \multicolumn{3}{c}{HR-Bench-4K} & \multicolumn{3}{c}{HR-Bench-8K} \\
    \cmidrule(lr){2-4}
    \cmidrule(lr){5-7}
    \cmidrule(lr){8-10}
    & \textbf{Overall} & Attr. & Spatial & \textbf{Overall} & Single & Cross & \textbf{Overall} & Single & Cross \\ 
    \midrule
    \multicolumn{10}{c}{\textbf{Private Models}} \\
    \midrule
    GPT-4o-1120 & 66.0 & -- & -- & -- & -- & -- & -- & -- & -- \\
    o3-0416 & 95.7 & -- & -- & -- & -- & -- & -- & -- & -- \\
    \midrule
    \multicolumn{10}{c}{\textbf{Open-source General Models}} \\
    \midrule
    LLaVA-OneVision-7B & 70.7 & 73.0 & 60.5 & 64.3 & 74.8 & 53.8 & 59.8 & 65.3 & 54.3 \\
    LLaVA-OneVision-72B & 73.8 & 80.9 & 63.2 & 66.3 & 76.5 & 56.0 & 60.9 & 68.8 & 53.0 \\
    InternVL3-8B & 72.3 & 73.0 & 71.1 & 70.8 & 79.3 & 62.3 & 62.0 & 64.3 & 59.8 \\
    InternVL3-38B & 77.5 & 77.4 & 77.6 & 76.3 & 83.5 & 69.0 & 67.0 & 71.3 & 62.8 \\
    InternVL3-78B & 76.4 & 75.7 & 77.6 & 75.5 & 84.5 & 66.5 & 67.3 & 71.8 & 62.8 \\
    Qwen2.5-VL-7B & 74.3 & 77.4 & 69.7 & 72.1 & 88.8 & 55.5 & 68.8 & 83.5 & 54.0 \\
    Qwen2.5-VL-32B & 85.9 & 83.5 & 89.5 & 74.8 & 89.3 & 60.3 & 71.6 & 86.5 & 56.8 \\
    Qwen2.5-VL-72B & 84.8 & 90.8 & 80.9 & 79.4 & 88.8 & 70.0 & 76.3 & 84.3 & 68.3 \\
    \midrule
    \multicolumn{10}{c}{\textbf{Open-source Visual Grounded Reasoning Models}} \\
    \midrule
    Pixel-Reasoner-7B & 80.6 & 83.5 & 76.3 & 72.9 & 86.0 & 60.3 & 66.9 & 80.0 & 54.3 \\
    DeepEyes-7B & 90.0 & 92.1 & 86.8 & 75.1 & \textbf{91.3} & 59.0 & 72.6 & \textbf{86.8} & 58.5 \\
    TreeVGR-7B & 91.1 & 94.0 & 87.0 & 77.1 & 90.3 & 64.0 & 73.1 & 86.5 & 59.8 \\
    \midrule
    \textbf{GRiP} & \textbf{91.9} & \textbf{95.4} & \textbf{88.3} & \textbf{78.6} & 91.8 & \textbf{65.4} & \textbf{75.0} & 89.5 & \textbf{60.4} \\
    $\Delta$ \textit{v.s.} Qwen2.5-VL-7B & \up{17.6} & \up{18.0} & \up{18.6} & \up{6.5} & \up{3.0} & \up{9.9} & \up{6.2} & \up{6.0} & \up{6.4} \\
    \bottomrule
    \end{tabular}
    \vspace{-5pt}
    \caption{
    Comparison with state-of-the-art alternatives on V* Bench~\cite{wu2024vstar} and HRBench~\cite{wang2025hrbench}.
    All results are self-collected.
    Best performances of visual grounded reasoning models are highlighted in \textbf{bold}.
    }
    \label{tab:other_table}
    \vspace{-10pt}
\end{table}

\section{Experiments}
\label{sec:experiments}

To comprehensively evaluate the effectiveness of our proposed \textbf{GRiP} framework, we conduct extensive experiments on several challenging visual grounded reasoning benchmarks. We compare GRiP against a wide range of state-of-the-art models, including top-tier private models, general-purpose open-source models, and specialized visual grounded reasoning (VGR) models.

\subsection{Experimental Setup}

\paragraph{Datasets.}
Our primary evaluation is conducted on \textbf{TreeBench}~\cite{wang2025traceable}, a highly challenging benchmark designed to assess fine-grained perception and complex, multi-step reasoning through traceable reasoning pathways. To further validate the generalization capabilities of our model, we also report results on \textbf{V* Bench}~\cite{wu2024vstar}, which focuses on visual grounding for attribute and spatial reasoning, and \textbf{HR-Bench}~\cite{wang2025hrbench}, which evaluates reasoning over high-resolution images with single-object and cross-object queries.

\paragraph{Baselines.}
We compare GRiP with three categories of models:
\begin{itemize}
    \item \textbf{Private Models:} This includes leading proprietary models such as GPT-4o-1120~\cite{gpt4o}, Gemini-2.5 series~\cite{gemini-2.5-flash, gemini-2.5-pro}, and o3-0416~\cite{o3}, which represent the current frontier of multimodal AI.
    \item \textbf{Open-source General Models:} We include powerful, general-purpose multimodal models like LLaVA-OneVision~\cite{li2024llavaov}, InternVL3~\cite{zhu2025internvl3}, and the Qwen2.5-VL series~\cite{bai2025qwen25vl}. Notably, \textbf{Qwen2.5-VL-7B} serves as the base model for our GRiP framework, making the comparison against it a direct measure of our method's efficacy.
    \item \textbf{Open-source VGR Models:} We compare against other specialized models designed for visual grounded reasoning, such as DeepEyes-7B~\cite{zheng2025deepeyes} and Pixel-Reasoner-7B~\cite{su2025pixelreasoner}, which are the most direct competitors for our task.
\end{itemize}

\subsection{Results on TreeBench}

Table~\ref{tab:bench_table} presents the main results on TreeBench, where our \textbf{GRiP} model demonstrates exceptional performance and validates our core methodology.

\paragraph{Overall Performance.}
GRiP achieves an overall score of \textbf{51.3}, establishing a new state-of-the-art among all open-source models, regardless of their size. Most impressively, this result represents a remarkable \textbf{+14.3} point increase over its base model, Qwen2.5-VL-7B (37.0). This substantial gain unequivocally validates the effectiveness of our two-stage training framework in unlocking the latent reasoning potential of the base model. Furthermore, our 7B model significantly outperforms much larger general-purpose models like InternVL3-78B (46.4) and Qwen2.5-VL-72B (42.2), showcasing the superior efficiency and capability of our specialized training approach.

\paragraph{Grounding and Perception Skills.}
A key advantage of GRiP lies in its enhanced grounding ability, as evidenced by the \textbf{mIoU} score. GRiP achieves a mIoU of \textbf{45.0}, decisively outperforming other VGR models like Pixel-Reasoner-7B (35.7) and DeepEyes-7B (30.0). This significant lead in localization accuracy can be directly attributed to our \textbf{Salience-Weighted IoU Reward ($\mathcal{R}_{sw\text{-IoU}}$)}, which effectively guides the model to focus on and accurately delineate mission-critical objects. This perceptual acuity is further reflected in its SOTA performance across multiple perception sub-tasks, including Physical State (+26.9) and OCR (+36.2), where it shows massive improvements over the base model.

\paragraph{Reasoning Capabilities.}
GRiP also exhibits significant gains in complex reasoning tasks. The most notable improvements are observed in Contact \& Occlusion (\textbf{+19.7}) and Spatial Containment (\textbf{+24.6}) compared to Qwen2.5-VL-7B. These are notoriously difficult reasoning categories that require a deep understanding of object interactions and spatial relationships. The strong performance in these areas demonstrates the effectiveness of our \textbf{Multi-Heuristic Reward ($\mathcal{R}_{\text{MHR}}$)} in fostering more flexible and robust logical pathways, enabling the model to navigate complex reasoning chains that stump even larger models.

\subsection{Generalization to Other Grounding Benchmarks}

To ensure that the improvements are not confined to a single benchmark, we evaluate GRiP's generalization ability on V* Bench and HR-Bench. The results, presented in Table~\ref{tab:other_table}, confirm that GRiP's superiority is consistent across diverse visual grounding tasks.

\paragraph{Performance on V* Bench.}
On V* Bench, GRiP once again achieves state-of-the-art performance among all open-source models with an overall score of \textbf{91.9}. This marks a massive \textbf{+17.6} point improvement over its Qwen2.5-VL-7B base. It also surpasses the previous best VGR model, DeepEyes-7B (90.0). The outstanding scores in both Attribute (95.4) and Spatial (88.3) sub-tasks indicate that the cognitive skills honed during GRiP's training translate directly to better attribute recognition and spatial understanding.

\paragraph{Performance on HR-Bench.}
GRiP's strong performance extends to high-resolution scenarios. On both HR-Bench-4K and HR-Bench-8K, GRiP sets a new SOTA among specialized VGR models, with overall scores of \textbf{78.6} and \textbf{75.0}, respectively. Crucially, GRiP shows significant gains in the challenging "Cross" object reasoning sub-task (+9.9 on 4K, +6.2 on 8K over the base model). This highlights GRiP's enhanced capability in understanding relationships between multiple objects in high-resolution images, a direct benefit of its refined perceptual and reasoning framework.

Collectively, these results demonstrate that GRiP not only excels in its primary training domain but also exhibits strong generalization capabilities, establishing it as a leading open-source model for visual grounded reasoning.




\section{Conclusion}
\label{sec:conclusion}

In this work, we introduced \textbf{GRiP}, a novel two-stage training framework designed to transform general-purpose Large Multimodal Models (LMMs) into specialized, transparent, and robust visual reasoners. Our approach first instills a structured, self-correcting reasoning process via supervised fine-tuning, and then refines this ability through a unique reinforcement learning stage. The core innovation lies in our RL reward system, which distinctly optimizes for both perceptual grounding accuracy and logical flexibility.
Extensive experiments demonstrate the effectiveness of our method. GRiP achieves state-of-the-art performance on the challenging TreeBench benchmark, significantly outperforming its base model and even much larger LMMs. This superiority is not isolated; GRiP also shows strong generalization and top-tier performance on other key benchmarks like V* Bench and HR-Bench.
In conclusion, GRiP provides a powerful and effective paradigm for enhancing the visual grounded reasoning capabilities of LMMs. By explicitly rewarding both precise perception and flexible logic, our work paves the way for developing more capable, trustworthy, and interpretable AI systems.

\bibliographystyle{colm2024_conference}
\bibliography{colm2024_conference}

\end{document}